\documentclass[letterpaper]{article}
\usepackage{aaai25}
\usepackage{times}
\usepackage{helvet}
\usepackage{courier}
\usepackage[hyphens]{url}
\usepackage{graphicx}
\usepackage{natbib}
\usepackage{caption}
\usepackage{amsmath}
\usepackage{amssymb}
\usepackage{enumitem}
\usepackage{multirow}
\usepackage[table,xcdraw]{xcolor}
\usepackage{algorithm}
\usepackage{algorithmic}

\title{IMA: An Imputation-based Mixup Augmentation Using Self-Supervised Learning for Time Series Data}
\author{
    Nha Dang Nguyen\textsuperscript{\rm 1,},
    Dang Hai Nguyen\textsuperscript{\rm 2},
    Khoa Tho Anh Nguyen\textsuperscript{\rm 3}\thanks{This research is supported by AI VIETNAM}
}
\affiliations{
    \textsuperscript{\rm 1}Vietnam - Korea University of Information and Communication Technology, Da Nang City, Vietnam\\
    \textsuperscript{\rm 2}VNU University of Engineering and Technology, Ha Noi City, Vietnam\\
    \textsuperscript{\rm 3}Vietnamese – German University, Ho Chi Minh City, Vietnam\\
    nhand.21it@vku.udn.vn, 24025015@vnu.edu.vn, 30421001@student.vgu.edu.vn
}

\begin{document}

\maketitle
\begin{abstract}
Data augmentation plays a crucial role in enhancing model performance across various AI fields by introducing variability while maintaining the underlying temporal patterns. However, in the context of long sequence time series data, where maintaining temporal consistency is critical, there are fewer augmentation strategies compared to fields such as image or text, with advanced techniques like Mixup rarely being used. In this work, we propose a new approach, Imputation-based Mixup Augmentation (IMA), which combines Imputed-data Augmentation with Mixup Augmentation to bolster model generalization and improve forecasting performance. We evaluate the effectiveness of this method across several forecasting models, including DLinear (MLP), TimesNet (CNN), and iTrainformer (Transformer), these models represent some of the most recent advances in long sequence time series forecasting. Our experiments, conducted on three datasets (ETT-small, Illness, Exchange Rate) from various domains and compared against eight other augmentation techniques, demonstrate that IMA consistently enhances performance, achieving 22 improvements out of 24 instances, with 10 of those being the best performances, particularly with iTrainformer imputation in ETT dataset. The GitHub repository is available at: https://github.com/dangnha/IMA.
\end{abstract}

\section{Introduction}
\begin{figure}[t]
\centering
\includegraphics[scale=0.25]{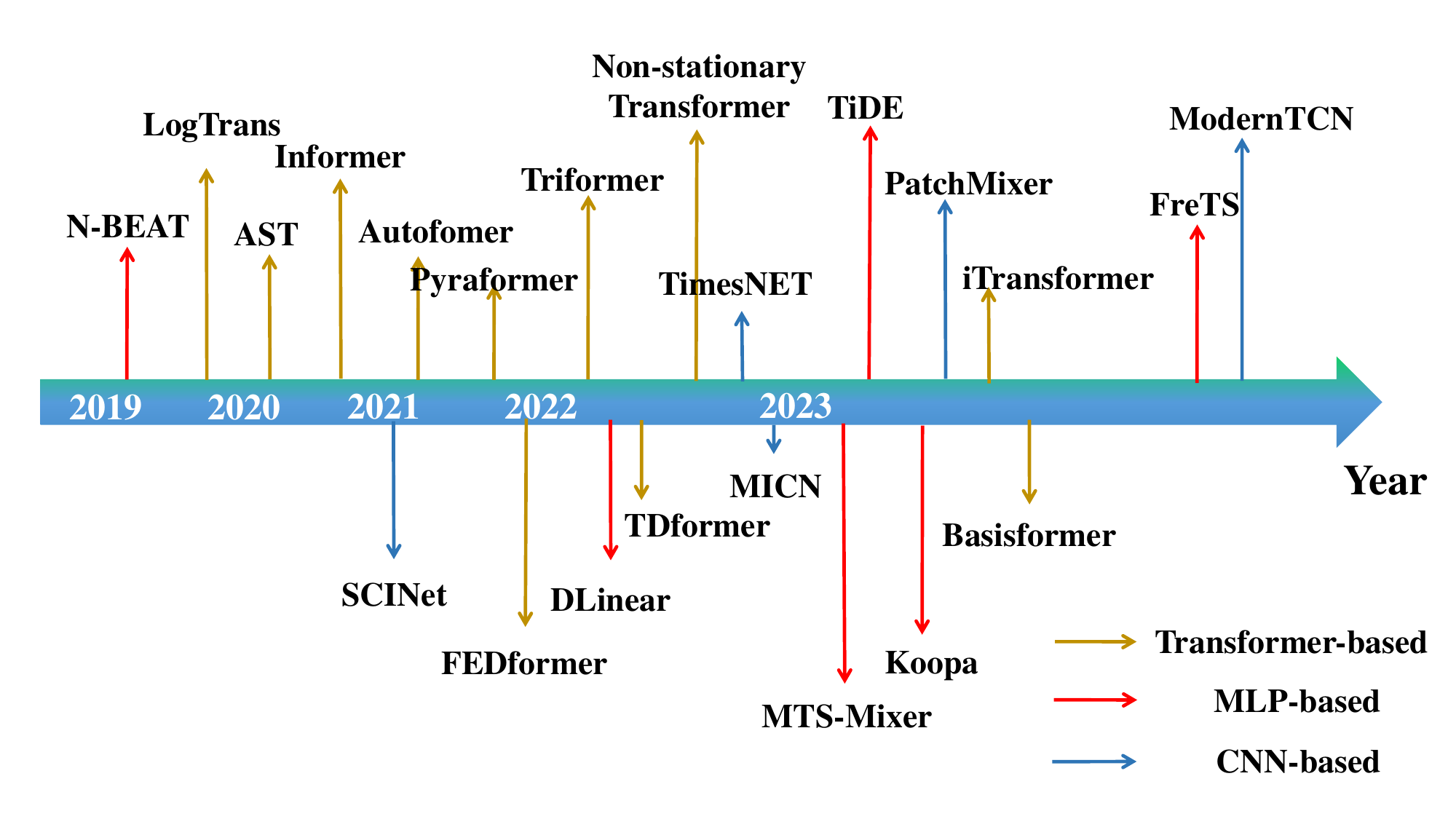}
\caption{Key Milestones in Time Series Forecasting \cite{math12101504}}
\label{fig1}
\end{figure}

Time series forecasting, particularly long sequence time-series forecasting (LSTF), plays a critical role in domains like finance, healthcare, energy, and urban planning \cite{CHEN2023101819}. Traditional statistical methods such as ARIMA and exponential smoothing laid the foundation but struggled with complex temporal dependencies. The advent of deep learning introduced RNNs \cite{ELMAN1990179}, LSTMs \cite{10.1162/neco.1997.9.8.1735}, and more recently, models like MLPs \cite{10.1145/3209978.3210006}, CNNs \cite{10.1145/3357384.3358132}, and Transformers \cite{wu2021autoformer, liu2024itransformer}, achieving state-of-the-art results in LSTF tasks (Fig. \ref{fig1}). These advances emphasize robust pipelines integrating preprocessing, feature extraction, and optimization.

Despite significant progress, challenges persist in time series data augmentation \cite{ijcai2021p631}. Unlike Computer Vision (CV) and Natural Language Processing (NLP), where augmentation techniques like flipping, cropping, and Mixup \cite{XU2023109347} are well-developed, time series techniques such as jittering and scaling often fail to capture complex temporal patterns \cite{Chen_Xu_Zeng_Xu_2023}. Emerging methods like latent Mixup show potential but remain underexplored. Similarly, while imputation methods like KNN and deep learning effectively handle missing data, their use for augmentation remains untapped, focusing solely on recovery rather than diversity enhancement.

This paper addresses these gaps by introducing Imputation-based Mixup Augmentation (IMA):
\begin{itemize}[label=\textbullet]
    \item We propose Imputated-data Augmentation with Self-Supervised Reconstruction (SSL), leveraging imputation for enriched data diversity.
    \item We develop IMA, combining imputation with Mixup to improve model generalization and performance.
    \item We evaluate IMA on three models—DLinear (MLP) \cite{10.1609/aaai.v37i9.26317}, TimesNet (CNN) \cite{wu2023timesnettemporal2dvariationmodeling}, and iTransformer (Transformer) \cite{liu2024itransformer}—demonstrating its effectiveness in enhancing forecasting performance across diverse scenarios.
\end{itemize}

\section{Related Work}
\textbf{Long sequence time-series forecasting (LSTF)} has seen rapid development due to its critical applications in various domains. Transformer-based models have significantly advanced the field by capturing long-term dependencies. Informer \cite{zhou2021informerefficienttransformerlong} and Autoformer \cite{wu2021autoformer} utilize sparse attention and series decomposition to reduce computational costs, while FEDformer \cite{zhou2022fedformerfrequencyenhanceddecomposed} incorporates frequency-domain techniques like Fourier and wavelet transformations for improved periodicity modeling. CNN-based models, including TCN \cite{BHARILYA2024100733} and SCINet \cite{liu2022scinet}, leverage dilated convolutions and hierarchical downsampling to capture both local and global patterns but face challenges in modeling long-term dependencies. RNNs \cite{ELMAN1990179}, such as LSTM \cite{10.1162/neco.1997.9.8.1735} and GRU \cite{cho2014learningphraserepresentationsusing}, have been enhanced through attention mechanisms (e.g., DA-RNN \cite{10.5555/3172077.3172254}) and hybrid models like ES-LSTM \cite{SMYL202075}, boosting their multivariate forecasting performance. MLP-based methods, while traditionally less suited for sequential data, have regained attention with feature-engineered adaptations, offering lightweight solutions for simpler tasks. These methods collectively highlight the progress in addressing scalability and efficiency challenges in LSTF (Fig. \ref{fig1}).

\textbf{Data augmentation} has emerged as a vital strategy for improving model performance, especially in scenarios with limited labeled data. Traditional time-domain transformations, such as window cropping, slicing, and warping, are widely used for their simplicity and ability to introduce variability \cite{ijcai2021p631}. Advanced techniques like decomposition-based methods (e.g., STL \cite{engproc2021005042}, Robust-STL \cite{RobustSTL}) and generative models like GANs and VAEs expand this toolkit by creating diverse yet structurally coherent synthetic data. Mixup, a method that interpolates between samples to generate new ones, remains underexplored for time series \cite{Zhou2023}, leaving significant room for further research.

\textbf{Imputation}, traditionally used to reconstruct missing data, has evolved with methods ranging from statistical interpolation to machine learning and deep learning techniques, including k-nearest neighbors (kNN), Gaussian Processes \cite{JAFRASTEH2023110603}, and Transformer-based models \cite{wang2024deeplearningmultivariatetime}. These approaches restore data while preserving temporal dependencies, suggesting potential for data augmentation. However, leveraging imputation explicitly for augmentation remains underexplored. This gap motivates our proposed Imputation-based Mixup Augmentation (IMA), detailed in the next section.
\section{Methodology}
Our approach consists of two main phases: \textbf{Self-Supervised Reconstruction (SSR)} and \textbf{Imputed-data Augmentation (IA)} with \textbf{Imputation-based Mixup Augmentation (IMA)}.

In the first phase, \textbf{Self-Supervised Reconstruction (SSR)}, an imputation model is trained to reconstruct masked input data, effectively capturing the intrinsic patterns and structures of the time series. This pre-training step allows the model to understand the temporal dependencies and complex features in the data.

In the second phase, the pre-trained imputation model is used for \textbf{Imputed-data Augmentation (IA)} to enhance data diversity by reconstructing masked sequences. Additionally, the augmented data is integrated with \textbf{Mixup Augmentation (IMA)}, which blends samples to introduce further variability in data representations. This combination improves model generalization and performance across various time series forecasting tasks.

% and adaptability
\begin{figure*}[h]
\centering
\includegraphics[width=\textwidth]{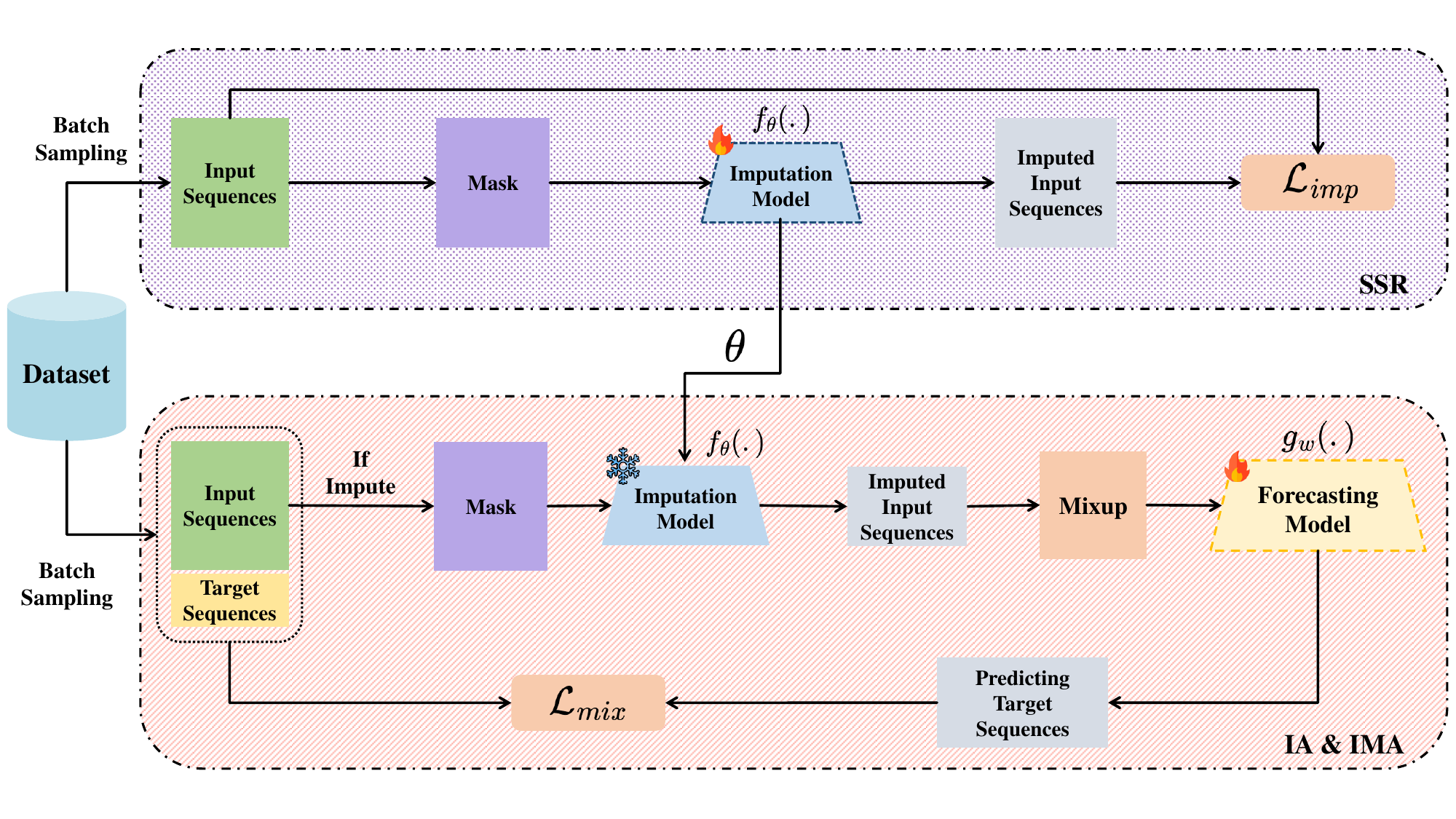}
\caption{Illustration of the proposed data augmentation framework, comprising two key phases: Self-Supervised Reconstruction (SSR) for learning intrinsic data patterns and Imputed-based Mixup Augmentation (IMA) for enhancing data diversity and model generalization.}
\label{fig2}
\end{figure*}

% =====================================================================
% \subsection{Annotation Definition}
% Let's consider a batch $\mathcal{B} = \{ \mathbf{X}^{(i)}, \mathbf{Y}^{(i)} \}_{i=1}^{|\mathcal{B}|}$ randomly sampled from original dataset $ \mathcal{D} = \{ (\mathbf{X}^{(i)}, \mathbf{Y}^{(i)}) \mid \mathbf{X}^{(i)} \in \mathbb{R} ^ {T_{\mathbf{X}} \times N_{\mathbf{X}}}, \mathbf{Y}^{(i)} \in \mathbb{R} ^ {T_{\mathbf{Y}} \times N_{\mathbf{Y}}} \; \text{for} \; i \in |\mathcal{D}| \} $ where $ \mathbf{X}^{(i)}$ represents the input sequences, and $ \mathbf{Y}^{(i)} $ represents their corresponding target sequences. Here, $T_{\mathbf{X}}, T_{\mathbf{Y}}$ represents the number of time steps in the sequences $\mathbf{X}$ and $\mathbf{Y}$, and $N_{\mathbf{X}}, N_{\mathbf{Y}}$ is the number of features per time step of $\mathbf{X}^{(i)}$, $\mathbf{Y}^{(i)}$, respectively. \\
% In detail, a batch will include $|\mathcal{B}|$ sample $\mathbf{X}^{(i)} = [\mathbf{X}^{(i)}_1, \ldots, \mathbf{X}^{(i)}_{T_{\mathbf{X}}}]$ represent a series for many time steps. Each time step will contain $N_{\mathbf{X}}$ features, defined: $\mathbf{X}^{(i)}_{T_{\mathbf{X}}} = [x_1, \ldots, x_{N_{\mathbf{X}}}]$ \\
% ====================================================================

\subsection{Annotation Definition}  
Let $\mathcal{B} = \{ (\mathbf{X}^{(i)}, \mathbf{Y}^{(i)}) \}_{i=1}^{|\mathcal{B}|}$ denote a batch of $|\mathcal{B}|$ samples randomly drawn from dataset $\mathcal{D}$, where:
\[
\mathcal{D} = \{ (\mathbf{X}^{(i)}, \mathbf{Y}^{(i)}) \mid \mathbf{X}^{(i)} \in \mathbb{R}^{T_{\mathbf{X}} \times N_{\mathbf{X}}}, \mathbf{Y}^{(i)} \in \mathbb{R}^{T_{\mathbf{Y}} \times N_{\mathbf{Y}}} \},
\]
with $i \in [1, |\mathcal{D}|]$. Here, $\mathbf{X}^{(i)}$ represents the input sequence, and $\mathbf{Y}^{(i)}$ is the corresponding target sequence. Parameters $T_{\mathbf{X}}$, $T_{\mathbf{Y}}$ denote the number of time steps, while $N_{\mathbf{X}}$, $N_{\mathbf{Y}}$ refer to features per time step in $\mathbf{X}$ and $\mathbf{Y}$.

Each input sequence $\mathbf{X}^{(i)}$ is represented as:
\[
\mathbf{X}^{(i)} = [\mathbf{X}^{(i)}_1, \mathbf{X}^{(i)}_2, \ldots, \mathbf{X}^{(i)}_{T_{\mathbf{X}}}],
\]
where $\mathbf{X}^{(i)}_t \in \mathbb{R}^{N_{\mathbf{X}}}$ for $t = 1, \ldots, T_{\mathbf{X}}$. Each time step $\mathbf{X}^{(i)}_t$ is defined as:
\[
\mathbf{X}^{(i)}_t = [x_1, x_2, \ldots, x_{N_{\mathbf{X}}}],
\]
with $x_j$ being the $j$-th feature at time step $t$.

\subsection{Self-Supervised Reconstruction (SSR)}
Self-supervised learning enhances downstream tasks by capturing inherent patterns within data. We apply this approach to time series imputation.

\begin{figure}[h]
\centering
\includegraphics[scale=0.26]{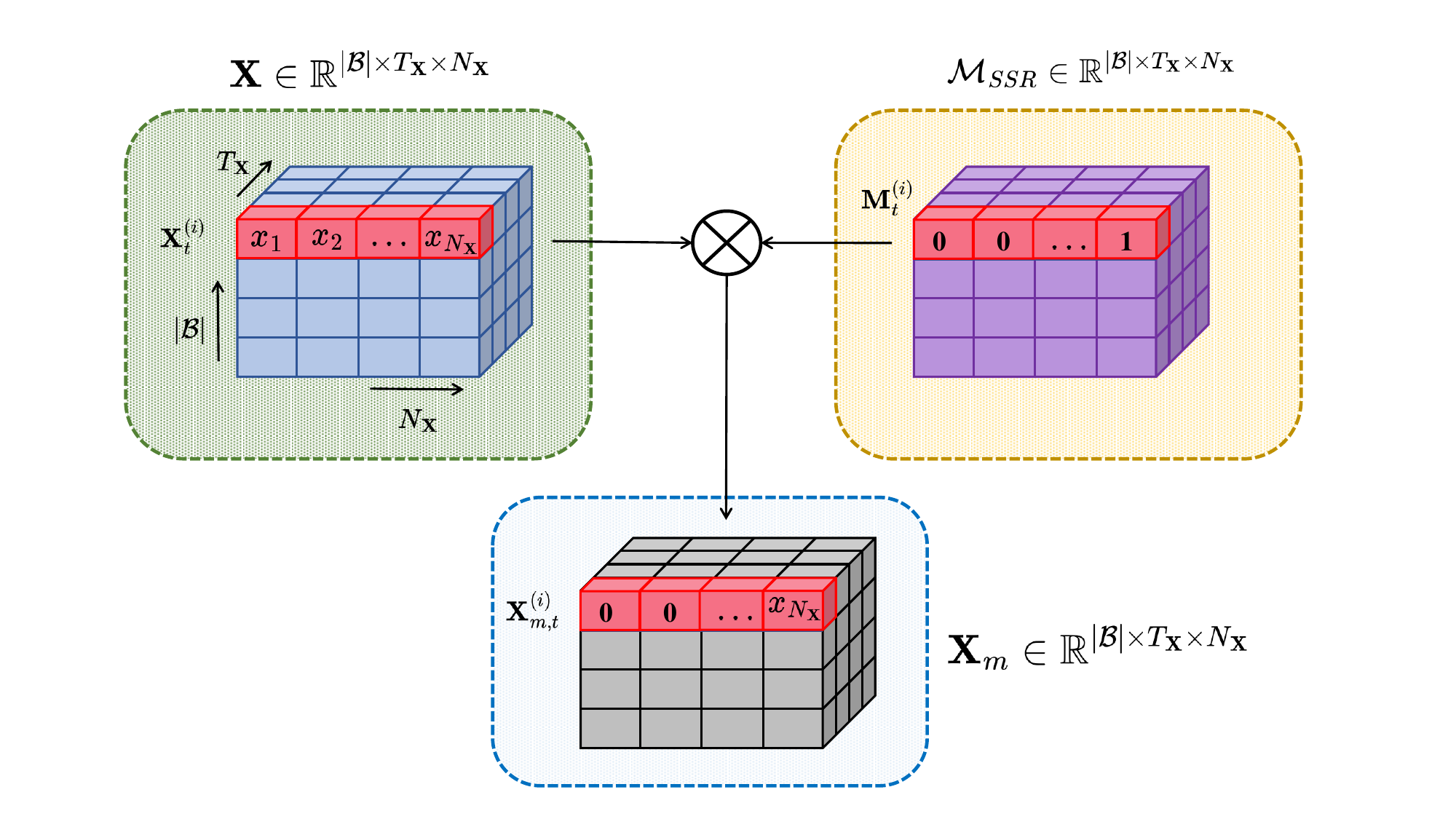}
\caption{Data masking strategy: applying a binary mask to generate masked inputs.}
\label{fig3}
\end{figure}

For each sample $\mathbf{X}^{(i)}$ in batch $\mathcal{B}$, masking is applied using $\mathcal{M}_{SSR} \in \mathbb{R}^{|\mathcal{B}| \times T_{\mathbf{X}} \times N_{\mathbf{X}}}$. The masked version $\mathbf{X}_m^{(i)}$ is defined as:
\[
\mathbf{X}_m^{(i)} = \{ \mathbf{X}_t^{(i)} \odot \mathbf{M}_t^{(i)} \mid t = 1, \ldots, T_{\mathbf{X}} \},
\]
where $\mathbf{X}_t^{(i)} \in \mathbb{R}^{N_{\mathbf{X}}}$ is the feature vector at time $t$, and $\mathbf{M}_t^{(i)} \in \{0, 1\}^{N_{\mathbf{X}}}$ is a binary mask vector. Mask $\mathbf{M}_t^{(i)}$ is constructed by randomly sampling values from a uniform distribution. Elements are set to 0 if the value is below the mask\_rate, indicating the feature is masked, or 1 otherwise, indicating the feature is observed (Fig. \ref{fig3}).

After getting $\{\mathbf{X}_m^{(i)}\}_{i=1}^{|\mathcal{B}|}$, the objective is to utilize an imputation model $f_{\theta}$ (where $\theta$ denotes the model parameters) to reconstruct the original input data $\{\mathbf{X}^{(i)}\}_{i=1}^{|\mathcal{B}|}$ from a masked version $\{\mathbf{X}_m^{(i)}\}_{i=1}^{|\mathcal{B}|}$. The model processes the masked data as input and generates imputed data as output $\mathbf{X}^{(i)}_{imp} = f_{\theta}(\mathbf{X}^{(i)}_m)$ (as showed in SSR phase of figure \ref{fig2}).
Finally, to guide the imputation model, MSE loss between original input sequence and imputed input sequence was applied:
\begin{equation}
    \mathcal{L}_{imp} \; = \; \frac{1}{|\mathcal{B}|} \sum_{i=1}^{|\mathcal{B}|} \sum_{t=1}^{T_{\mathbf{X}}} \; (1 - \mathbf{M}_t^{(i)}) \cdot \left( \mathbf{X}_t^{(i)} - \mathbf{X}_{imp, t}^{(i)} \right)^2 
\end{equation}
\subsection{Imputed-data Augmentation (IA)}
After sampling batches, a binary vector $\mathbf{i} \in \mathbb{R}^B$ is defined, where $B$ is the number of batches. Each element $\mathbf{i}^{(i)}$ is determined by comparing a random number (from a uniform distribution) with the imputation\_rate. If the random number is less than the imputation\_rate, $\mathbf{i}^{(i)} = 1$, indicating imputation-based augmentation for the batch. Otherwise, $\mathbf{i}^{(i)} = 0$, and no augmentation is applied.

Following the \textbf{Self-Supervised Reconstruction (SSR)} phase, the pre-trained model \( f_{\theta} \), which has learned the temporal patterns and structures of the data, reconstructs masked sequences \( \mathbf{X}_m^{(i)} \) in each batch \( \mathcal{B} \). Using the binary mask matrix \( \mathcal{M}_{IMA} \in \mathbb{R}^{|\mathcal{B}| \times T_{\mathbf{X}} \times N_{\mathbf{X}}} \), the imputed sequence is generated as:
\[
\mathbf{X}_{\text{imp}}^{(i)} = f_{\theta}(\mathbf{X}_m^{(i)}).
\]

The reconstructed sequences form the augmented batch \( \mathcal{B}^{\mathbf{X}}_{\text{imp}} = \{ \mathbf{X}_{\text{imp}}^{(i)} \}_{i=1}^{|\mathcal{B}|} \). These sequences are then passed into a forecasting model \( g_{w}(\cdot) \), parameterized by \( w \), to predict target sequences:
\[
\hat{\mathbf{Y}}^{(i)} = g_{w}(\mathbf{X}_{\text{imp}}^{(i)}).
\]

This process mitigates biases by imputing missing values with plausible estimates, thereby increasing diversity while maintaining the original data's structure and patterns.

% ================================================================
\subsection{Imputed-based Mixup Augmentation}
After generating the imputed batch \( \mathcal{B}^{\mathbf{X}}_{\text{imp}} \), Mixup Augmentation is applied to create synthetic data and enhance model generalization. Mixup interpolates between pairs of samples within \( \mathcal{B}_{\text{imp}} \), governed by a mixing coefficient \( \lambda \sim \text{Beta}(\alpha, \alpha) \), where \( \lambda \in [0, 1] \). This coefficient determines the contribution of each sample in the interpolation.

For two randomly selected imputed samples \( \mathbf{X}_{\text{imp}}^{(i)} \) and \( \mathbf{X}_{\text{imp}}^{(j)} \), the mixed input \( \mathbf{X}_{\text{mix}}^{(i,j)} \) is computed as:
\[
\mathbf{X}_{\text{mix}}^{(i,j)} = \lambda \cdot \mathbf{X}_{\text{imp}}^{(i)} + (1 - \lambda) \cdot \mathbf{X}_{\text{imp}}^{(j)}.
\]
The mixed sample is passed to the forecasting model \( g_{w} \), and the loss for the mixed sample is calculated as:
\[
\mathcal{L}_{\text{mix}} = \lambda \cdot \mathcal{L}(g_{w}(\mathbf{X}_{\text{mix}}^{(i,j)}), \mathbf{Y}^{(i)}) + (1 - \lambda) \cdot \mathcal{L}(g_{w}(\mathbf{X}_{\text{mix}}^{(i,j)}), \mathbf{Y}^{(j)}),
\]
where \( \mathcal{L} \) denotes the forecasting loss, and \( \mathbf{Y}^{(i)} \), \( \mathbf{Y}^{(j)} \) are the target sequences corresponding to the original samples.

\begin{figure}[h]
\centering
\includegraphics[scale=0.26]{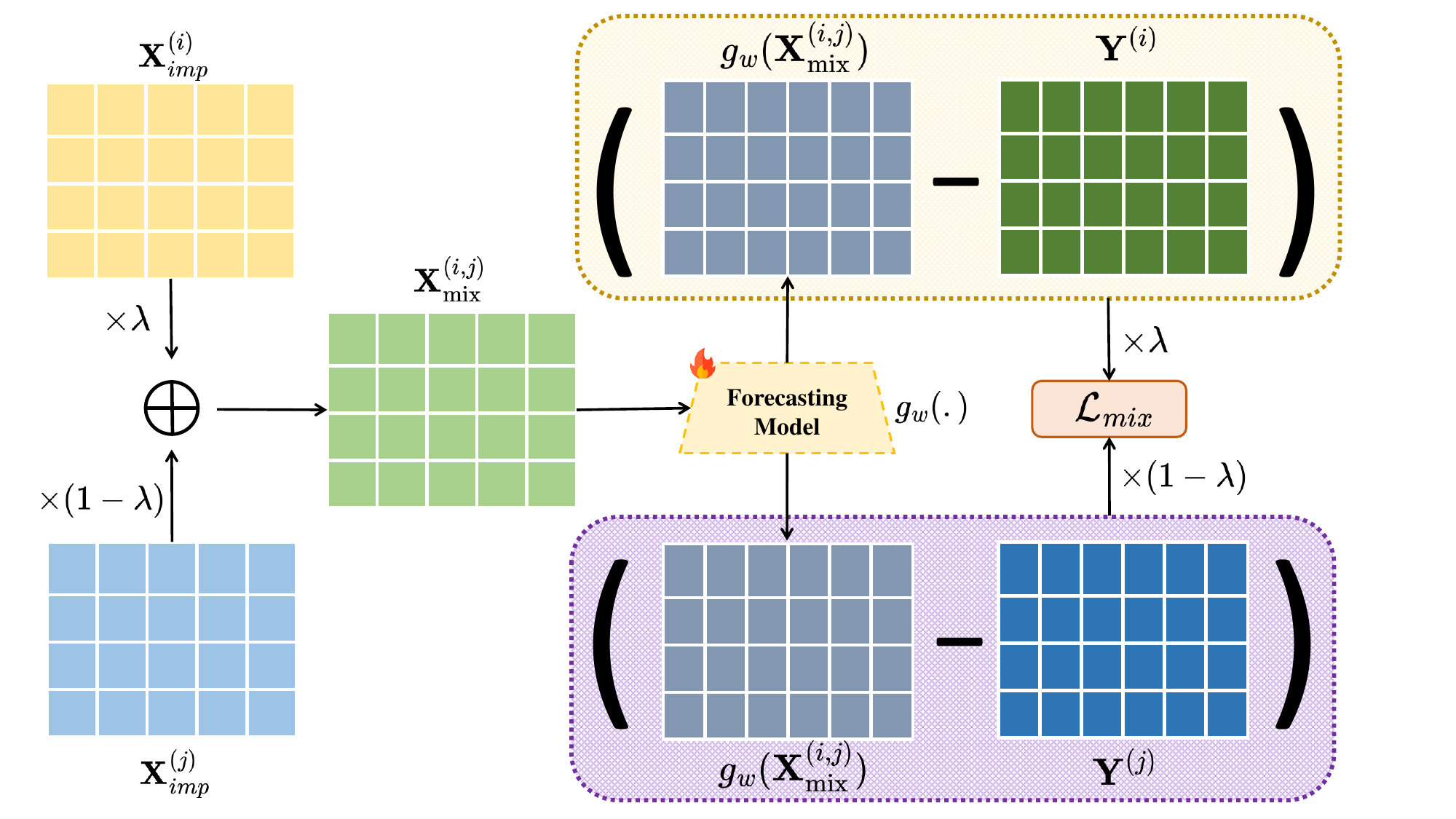}
\caption{Mixup applied to two imputed samples.}
\label{fig4}
\end{figure}

\begin{algorithm}[tb]
\caption{Imputed-based Mixup Augmentation}
\label{alg:algorithm}
\begin{algorithmic}[1]
\STATE \textbf{Input:} Batch \( \mathcal{B} = \{ (\mathbf{X}^{(i)}, \mathbf{Y}^{(i)}) \}_{i=1}^{|\mathcal{B}|} \), imputation\_rate, mask\_rate
\STATE Apply SSR to compute \( \mathbf{X}_{\text{imp}}^{(i)} \) for each \( \mathbf{X}_m^{(i)} \) in \( \mathcal{B} \)
\STATE Shuffle \( \mathcal{B}_{\text{imp}} \) to create pairs \( (\mathbf{X}_{\text{imp}}^{(i)}, \mathbf{X}_{\text{imp}}^{(j)}) \)
\STATE Compute mixed samples \( \mathbf{X}_{\text{mix}}^{(i,j)} \) using \( \lambda \sim \text{Beta}(\alpha, \alpha) \)
\STATE Compute loss \( \mathcal{L}_{\text{mix}} \) and update \( g_{w} \) via gradient descent
\end{algorithmic}
\end{algorithm}

% ==================================================
\section{Experiments and Results}
\textbf{Dataset.}  
We evaluate our data augmentation method using three long-term time series forecasting datasets: \textbf{ETT-small}, \textbf{Illness}, and \textbf{Exchange Rate}.  
\textbf{ETT-small} \cite{zhou2021informerefficienttransformerlong} includes two subsets (ETTh and ETTm) tracking transformer station temperatures at hourly and 15-minute intervals, with 70,080 samples. Each sample contains six features and an Oil Temperature target, capturing seasonal and irregular patterns.  
\textbf{Illness}\footnote{Illness dataset: Weekly ILI rates across U.S. regions, provided by the CDC via FluView portal. Accessed December 1, 2024, gis.cdc.gov/grasp/fluview/fluportaldashboard.html.} records weekly influenza-like illness (ILI) rates along with features like population and healthcare capacity.  
\textbf{Exchange Rate} \cite{Lai2017ModelingLA} tracks daily exchange rates for eight currencies (e.g., USD, GBP, AUD) from 1990 to 2016, comprising 7,588 time steps, offering insights into long-term financial forecasting and cross-variable correlations.

\begin{table*}[ht]
\centering
\tiny   
\begin{tabular}{|c|c|c|c|c|c|c|c|c|c|c|c|c|}
\hline
\textbf{Model} & \multicolumn{2}{c|}{\textbf{DLinear}} & \multicolumn{2}{c|}{\textbf{TimesNet}} & \multicolumn{2}{c|}{\textbf{iTransformer}} & \multicolumn{2}{c|}{\textbf{DLinear}} & \multicolumn{2}{c|}{\textbf{TimesNet}} & \multicolumn{2}{c|}{\textbf{iTransformer}} \\
\hline
\textbf{Metric} & \textbf{MSE} & \textbf{MAE} & \textbf{MSE} & \textbf{MAE} & \textbf{MSE} & \textbf{MAE} & \textbf{MSE} & \textbf{MAE} & \textbf{MSE} & \textbf{MAE} & \textbf{MSE} & \textbf{MAE} \\
\hline
% =======================================================
\rowcolor{blue!30} \textbf{Dataset} & \multicolumn{6}{c|}{\textbf{ETTh1}} & \multicolumn{6}{c|}{\textbf{ETTh2}} \\
\hline
\rowcolor{yellow!15}Baseline & 0.445051  & 0.440448 & 0.458988 & 0.45493 & 0.447285 & 0.440212 & 0.479687 & 0.477797 & 0.406138 & 0.413908 & 0.4578 & 0.398759  \\
\hline
\rowcolor{yellow!15}Jitter & 0.00E+00 & 1.00E-06 & \textcolor{blue}{-1.37E-03} & \textcolor{blue}{-7.73E-04} & \textcolor{blue}{-1.10E-05} & 7.06E-03 & 3.00E-05 & 2.20E-05 & 3.99E-03 & 3.63E-03 & \textcolor{blue}{-7.84E-02} &  \textcolor{blue}{-1.13E-04} \\
\hline
\rowcolor{yellow!15}Hflip & \textcolor{blue}{-5.73E-04} & \textcolor{blue}{-2.07E-04} & 7.94E-03 & 2.84E-03 & \textcolor{blue}{-5.47E-03} & \textcolor{blue}{-4.07E-03} & 1.24E-03 & 1.22E-03 & \textbf{\textcolor{red}{-1.63E-02}} & \textbf{\textcolor{red}{-7.31E-03}} & \textcolor{blue}{-5.92E-02} & \textbf{\textcolor{red}{-2.66E-01}}  \\
\hline
\rowcolor{yellow!15}Vflip & 4.41E-03 & 4.89E-03 & \textcolor{blue}{-6.30E-03} & \textcolor{blue}{-2.93E-03} & \textcolor{blue}{-5.86E-03} & \textcolor{blue}{-4.96E-03} & 7.55E-03 & 6.83E-03 & 3.36E-02 & 1.29E-02 & \textcolor{blue}{-7.43E-02} & \textcolor{blue}{-7.35E-04}  \\
\hline
\rowcolor{yellow!15}Scaling & 7.93E-04 & 1.14E-03 & \textcolor{blue}{-1.34E-03} & \textcolor{blue}{-6.42E-04} & \textbf{\textcolor{red}{-2.98E-01}} & \textbf{\textcolor{red}{-2.93E-01}} & \textcolor{blue}{-6.82E-04} & 1.97E-04 & 2.07E-02 & 1.29E-02 & \textcolor{blue}{-7.80E-02} & \textcolor{blue}{-2.83E-04}  \\
\hline
\rowcolor{yellow!15}Win\_warp & 1.81E-02 & 1.81E-02 & 3.71E-02 & 2.06E-02 & 2.21E-02 & 1.30E-02 & \textcolor{blue}{-6.41E-04} & 9.71E-04 & \textcolor{blue}{-1.16E-04} & 3.98E-03 & \textcolor{blue}{-7.46E-02} & 2.40E-03  \\
\hline
\rowcolor{yellow!15}Win\_slide & 6.66E-02 & 4.17E-02 & 1.65E-02 & 1.50E-02 & 2.79E-02 & 1.76E-02 & 7.63E-03 & 7.55E-03 & 1.48E-02 & 1.23E-02 & \textcolor{blue}{-6.57E-02} & 9.67E-03  \\
\hline
\rowcolor{yellow!15}Permu & \textcolor{blue}{-1.66E-04} & 4.87E-04 & \textcolor{blue}{-3.60E-03} & 1.44E-03 & 1.05E-02 & 1.76E-02 & 8.17E-03 & 6.42E-03 & 1.21E-02 & 5.43E-03 & \textcolor{blue}{-7.62E-02} & 1.34E-04  \\
\hline
\rowcolor{yellow!15}Mixup & 2.10E-04 & 1.92E-04 & 7.78E-04 & \textcolor{blue}{-4.23E-04} & \textcolor{blue}{-1.58E-03} & \textcolor{blue}{-1.38E-03} & \textcolor{blue}{-1.01E-02} & \textcolor{blue}{-5.84E-03} & \textcolor{blue}{-2.67E-03} & \textcolor{blue}{-1.06E-03} & \textbf{\textcolor{red}{-7.88E-02}} & \textcolor{blue}{-3.87E-04}   \\
\hline
\rowcolor{green!15}TS\_IA & \textcolor{blue}{-4.54E-03} & \textcolor{blue}{-3.23E-03} & \textcolor{blue}{-1.41E-04} & \textcolor{blue}{-1.48E-03} & \textcolor{blue}{-1.78E-03} & \textcolor{blue}{-1.89E-03} & \textcolor{blue}{-1.87E-02} & \textcolor{blue}{-1.20E-02} & 1.17E-03 & 1.02E-04 & \textcolor{blue}{-7.74E-02} & \textcolor{blue}{-2.00E-03}  \\
\hline
\rowcolor{green!15}iT\_IA & \textcolor{blue}{-5.73E-03} & \textcolor{blue}{-4.84E-03} & \textcolor{blue}{-7.39E-03} & \textcolor{blue}{-7.65E-03} & \textcolor{blue}{-1.33E-03} & \textcolor{blue}{-1.96E-03}  & \textcolor{blue}{-1.78E-02} & \textcolor{blue}{-1.16E-02} & \textcolor{blue}{-7.39E-03} & \textcolor{blue}{-3.62E-03} & \textcolor{blue}{-7.49E-02} & \textcolor{blue}{-3.09E-04} \\
\hline
\rowcolor{green!15}TS\_IMA & \textcolor{blue}{-6.31E-03} & \textbf{\textcolor{red}{-6.11E-03}} & \textcolor{blue}{-5.13E-03} & \textcolor{blue}{-6.76E-03} & \textcolor{blue}{-3.68E-03} & \textcolor{blue}{-3.60E-03} & \textbf{\textcolor{red}{-2.15E-02}} & \textbf{\textcolor{red}{-1.30E-02}} & \textcolor{blue}{-1.28E-02} & \textcolor{blue}{-7.06E-03} & \textcolor{blue}{-7.65E-02} & \textcolor{blue}{-1.84E-03}  \\
\hline
\rowcolor{green!15}iT\_IMA & \textbf{\textcolor{red}{-6.38E-03}} & \textcolor{blue}{-5.85E-03} & \textbf{\textcolor{red}{-1.54E-02}} & \textbf{\textcolor{red}{-1.18E-02}} & \textcolor{blue}{-3.14E-03} & \textcolor{blue}{-3.98E-03} & \textcolor{blue}{-1.25E-02} & \textcolor{blue}{-8.32E-03} & \textcolor{blue}{-1.18E-02} & \textcolor{blue}{-6.75E-03} & \textcolor{blue}{-7.52E-02} & \textcolor{blue}{-2.02E-04}  \\
\hline
% ======================================================
\rowcolor{blue!30}\textbf{Dataset} & \multicolumn{6}{c|}{\textbf{ETTm1}} & \multicolumn{6}{c|}{\textbf{ETTm2}} \\
\hline
\rowcolor{yellow!15}Baseline & 0.381687 & 0.390652 & 0.39177 & 0.403024 & 0.398893 & 0.394252 & 0.281894 & 0.358602 & 0.2544 & 0.307104 & 0.252632 & 0.312604  \\
\hline
\rowcolor{yellow!15}Jitter & 1.50E-05 & \textbf{\textcolor{red}{-8.95E-03}} & 1.39E-04 & 4.12E-04 & \textcolor{blue}{-4.64E-03} & \textcolor{blue}{-4.60E-05} & 1.40E-05 & 1.60E-05 & 6.46E-03 & 3.53E-03 & \textcolor{blue}{-1.81E-04} &  \textcolor{blue}{-2.21E-04} \\
\hline
\rowcolor{yellow!15}Hflip & \textbf{\textcolor{red}{-8.20E-05}} & \textcolor{blue}{-2.56E-04} & 3.84E-03 & 2.34E-04 & \textcolor{blue}{-1.65E-02} & \textcolor{blue}{-2.11E-03} & 2.81E-03 & 3.15E-03 & \textcolor{blue}{-2.67E-03} & 1.20E-03 & \textbf{\textcolor{red}{-4.76E-03}} & \textcolor{blue}{-3.79E-03}  \\
\hline
\rowcolor{yellow!15}Vflip & 5.25E-04 & 1.00E-03 & \textcolor{blue}{-2.07E-03} & \textcolor{blue}{-4.10E-03} & \textcolor{blue}{-6.76E-03} & 3.11E-03 & \textcolor{blue}{-5.25E-03} & \textcolor{blue}{-2.63E-03} & \textcolor{blue}{-2.69E-03} & 1.28E-03 & 5.62E-02 & \textcolor{blue}{-3.79E-03}  \\
\hline
\rowcolor{yellow!15}Scaling & 3.79E-04 & 7.50E-04 & \textcolor{blue}{-2.01E-03} & \textcolor{blue}{-1.02E-03} & \textcolor{blue}{-1.80E-02} & 5.00E-05 & 2.25E-03 & 2.33E-03 & 7.54E-03 & 4.58E-03 & \textcolor{blue}{-4.20E-05} & 5.30E-05  \\
\hline
\rowcolor{yellow!15}Win\_warp & 7.35E-02 & 4.84E-02 & 4.79E-02 & 3.62E-02 & 4.49E-02 & 3.92E-02 & 3.46E-03 & 4.96E-03 & 3.92E-03 & 5.27E-03 & 2.42E-03 & 3.48E-03  \\
\hline
\rowcolor{yellow!15}Win\_slide  & 8.62E-02 & 5.08E-02 & 8.04E-02 & 4.75E-02 & 3.45E-02 &  3.92E-02 & 1.50E-03 & 2.18E-03 & 1.33E-02 & 1.04E-02 & 4.73E-03 &  4.28E-03 \\
\hline
\rowcolor{yellow!15}Permu & 4.45E-02 & 3.36E-02 & 7.88E-03 & 1.13E-02 & 3.45E-02 & 1.95E-02 & 5.75E-03 & 7.47E-03 & 1.33E-02 & 2.23E-03 & 1.93E-03 & 2.91E-03  \\
\hline
\rowcolor{yellow!15}Mixup & 1.54E-04 & 4.21E-04 & \textcolor{blue}{-2.86E-03} & \textcolor{blue}{-2.19E-03} & \textcolor{blue}{-1.90E-02} & \textcolor{blue}{-2.12E-03} & \textcolor{blue}{-2.09E-03} & \textcolor{blue}{-1.48E-03} & \textcolor{blue}{-1.14E-03} & \textcolor{blue}{-9.32E-04} & \textcolor{blue}{-1.56E-03} & \textcolor{blue}{-1.10E-03} \\
\hline
\rowcolor{green!15}TS\_IA & 4.96E-03 & 4.27E-03 & \textcolor{blue}{-8.28E-03} & \textcolor{blue}{-2.50E-03} & \textcolor{blue}{-1.99E-02} & \textcolor{blue}{-1.15E-03} & \textbf{\textcolor{red}{-2.18E-02}} & \textbf{\textcolor{red}{-2.18E-02}} & \textcolor{blue}{-4.66E-03} & \textcolor{blue}{-3.04E-03} & \textcolor{blue}{-2.82E-03} & \textcolor{blue}{-3.71E-03}  \\
\hline
\rowcolor{green!15}iT\_IA & 5.44E-03 & 4.72E-03 & \textcolor{blue}{-7.23E-03} & \textcolor{blue}{-3.52E-03} & \textcolor{blue}{-2.12E-02} &  \textcolor{blue}{-1.20E-03}  & 2.06E-03 & 1.87E-03 & \textcolor{blue}{-4.39E-03} & \textcolor{blue}{-2.77E-03} & \textcolor{blue}{-2.74E-03} &  \textcolor{blue}{-3.71E-03} \\
\hline
\rowcolor{green!15}TS\_IMA & 5.18E-03 & 4.27E-03 & \textcolor{blue}{-7.62E-03} & \textcolor{blue}{-4.76E-03} & \textcolor{blue}{-2.17E-02} &  \textcolor{blue}{-3.63E-03} & \textcolor{blue}{-1.49E-02} & \textcolor{blue}{-1.61E-02} & \textcolor{blue}{-5.80E-03} & \textcolor{blue}{-2.50E-03} & \textcolor{blue}{-4.03E-03} &  \textcolor{blue}{-4.44E-03} \\
\hline
\rowcolor{green!15}iT\_IMA & 6.27E-03 & 5.27E-03 & \textbf{\textcolor{red}{-1.22E-02}} & \textbf{\textcolor{red}{-9.31E-03}} & \textbf{\textcolor{red}{-2.36E-02}} & \textbf{\textcolor{red}{-4.22E-03}} & \textcolor{blue}{-6.69E-03} & \textcolor{blue}{-5.99E-03} & \textbf{\textcolor{red}{-7.18E-03}} & \textbf{\textcolor{red}{-4.58E-03}} & \textcolor{blue}{-4.34E-03} & \textbf{\textcolor{red}{-5.13E-03}}  \\
\hline
% =================================================================
\rowcolor{blue!30}\textbf{Dataset} & \multicolumn{6}{c|}{\textbf{Illness}} & \multicolumn{6}{c|}{\textbf{Exchange Rate}} \\
\hline
\rowcolor{yellow!15}Baseline & 4.003106 & 1.441318 & 1.998755 & 0.885458 & 1.807093 & 0.870089 & 0.168927 & 0.305395 & 0.219712 & 0.340417 & 0.180669 & 0.303503  \\
\hline
\rowcolor{yellow!15}Jitter & 0.00E+00 & 0.00E+00 & \textcolor{blue}{-1.90E-05} & \textcolor{blue}{-3.00E-06} & 0.00E+00 & 0.00E+00 & 0.00E+00 & 0.00E+00 & \textcolor{blue}{-6.93E-04} & \textcolor{blue}{-3.37E-04} & 0.00E+00 & 0.00E+00  \\
\hline
\rowcolor{yellow!15}Hflip & 0.00E+00 & 0.00E+00 & 3.30E-05 & 7.00E-06 & 0.00E+00 & 0.00E+00 & 0.00E+00 & 0.00E+00 & \textcolor{blue}{-3.72E-04} & \textcolor{blue}{-2.52E-04} & 0.00E+00 &  0.00E+00 \\
\hline
\rowcolor{yellow!15}Vflip & 0.00E+00 & 0.00E+00 & \textcolor{blue}{-8.00E-06} & 3.00E-06 & 0.00E+00 & 0.00E+00 & 0.00E+00 & 0.00E+00 & \textcolor{blue}{-1.44E-03} & \textcolor{blue}{-1.29E-03} & 0.00E+00 &  0.00E+00 \\
\hline
\rowcolor{yellow!15}Scaling & 0.00E+00 & 0.00E+00 & 1.10E-05 & 3.00E-06 & 0.00E+00 & 0.00E+00 & 0.00E+00 & 0.00E+00 & \textcolor{blue}{-1.14E-03} & \textcolor{blue}{-5.71E-04} & 0.00E+00 & 0.00E+00  \\
\hline
\rowcolor{yellow!15}Win\_warp & 0.00E+00 & 0.00E+00 & 1.90E-05 & 2.00E-06 & 0.00E+00 & 0.00E+00 & 0.00E+00 & 0.00E+00 & \textcolor{blue}{-1.90E-03} & \textcolor{blue}{-7.10E-04} & 0.00E+00 & 0.00E+00  \\
\hline
\rowcolor{yellow!15}Win\_slide & 0.00E+00 & 0.00E+00 & 1.30E-05 & 1.00E-06 & 0.00E+00 & 0.00E+00 & 0.00E+00 & 0.00E+00 & \textcolor{blue}{-8.52E-04} & \textcolor{blue}{-6.02E-04} & 0.00E+00 & 0.00E+00  \\
\hline
\rowcolor{yellow!15}Permu & 0.00E+00 & 0.00E+00 & \textcolor{blue}{-2.30E-05} & \textcolor{blue}{-2.00E-06} & 0.00E+00 & 0.00E+00 & 0.00E+00 & 0.00E+00 & \textcolor{blue}{-3.70E-05} & 4.92E-04 & 0.00E+00 & 0.00E+00  \\
\hline
\rowcolor{yellow!15}Mixup & 2.45E-02 & 4.30E-03 & \textcolor{blue}{-1.06E-01}  & \textbf{\textcolor{red}{-2.64E-02}} & 3.93E-02 & 3.99E-03 & 7.51E-03 & 8.34E-03 & \textcolor{blue}{-8.73E-03} & \textcolor{blue}{-9.11E-03} & \textcolor{blue}{-1.08E-03} & \textcolor{blue}{-1.46E-03}  \\
\hline
\rowcolor{green!15}TS\_IA & \textcolor{blue}{-9.15E-03} & \textcolor{blue}{-7.45E-04} & \textcolor{blue}{-7.46E-02} & \textcolor{blue}{-2.10E-02} & 3.21E-02 & 6.48E-03 & 4.18E-03 & 6.24E-03 & \textbf{\textcolor{red}{-1.65E-02}} & \textbf{\textcolor{red}{-1.59E-02}} & \textbf{\textcolor{red}{-1.90E-03}} & \textbf{\textcolor{red}{-2.32E-03}}  \\
\hline
\rowcolor{green!15}iT\_IA & \textbf{\textcolor{red}{-9.85E-03}} & \textbf{\textcolor{red}{-9.53E-04}} & \textcolor{blue}{-7.03E-02} & \textcolor{blue}{-2.40E-02} & 4.09E-02 & 3.00E-03 & 7.16E-03 & 8.38E-03 & 4.83E-03 & 1.75E-03 & \textcolor{blue}{-1.46E-03} & \textcolor{blue}{-1.51E-03}  \\
\hline
\rowcolor{green!15}TS\_IMA & \textcolor{blue}{-1.92E-03} & 2.27E-04 & \textcolor{blue}{-2.67E-03} & \textcolor{blue}{-1.23E-02} & 1.04E-01 & 2.12E-02 & 3.56E-03 & 5.04E-03 & \textcolor{blue}{-8.48E-03} & \textcolor{blue}{-9.61E-03} & \textcolor{blue}{-1.84E-03} & \textcolor{blue}{-2.12E-03}  \\
\hline
\rowcolor{green!15}iT\_IMA & 1.60E-03 & 1.74E-03 & \textbf{\textcolor{red}{-1.43E-01}} & \textcolor{blue}{-2.54E-02} & 7.62E-02 & 1.13E-02 & 7.68E-03 & 1.00E-02 & \textcolor{blue}{-1.54E-02} & \textcolor{blue}{-1.48E-02} & \textcolor{blue}{-1.49E-03} & \textcolor{blue}{-1.89E-03}  \\
\hline
% ============================================================
\end{tabular}
\caption{Forecasting Performance Evaluation. Comparison of 8 augmentation methods with IM and IMA, using TimesNet (TS) and iTransformer (iT) for imputation-based enhancement.   \textbf{Red bold}: best case, \textbf{Blue}: improvement case, \textbf{Green Background}: Our methods.}
\label{tab:result}
\end{table*}
\textbf{Experimental Setting.}  
We conducted our experiments using the TSLib framework \cite{wu2023timesnettemporal2dvariationmodeling, wang2024deeptimeseriesmodels}, evaluating three baseline models—DLinear, TimesNet, and iTransformer—representing key approaches in time series modeling: MLP, CNN, and Transformer, respectively. Seven widely used data augmentation techniques (Jitter, Horizontal Flip, Vertical Flip, Scaling, Window Warp, Window Slide, and Permutation) and the Mixup method were applied. 

Our proposed IMA method optimized the imputation\_rate (0.125 for TimesNet and iTransformer) and mask\_rate (0.375 for TimesNet, 0.125 for iTransformer) through grid search. DLinear was excluded from the imputation task due to its inability to capture complex temporal patterns, as evidenced by consistent underperformance in preliminary tests. 

Model performance was evaluated using Mean Squared Error (MSE) and Mean Absolute Error (MAE) to assess both prediction accuracy and robustness across datasets and augmentation strategies.

\begin{figure}[ht]
\centering
\includegraphics[scale=0.248]{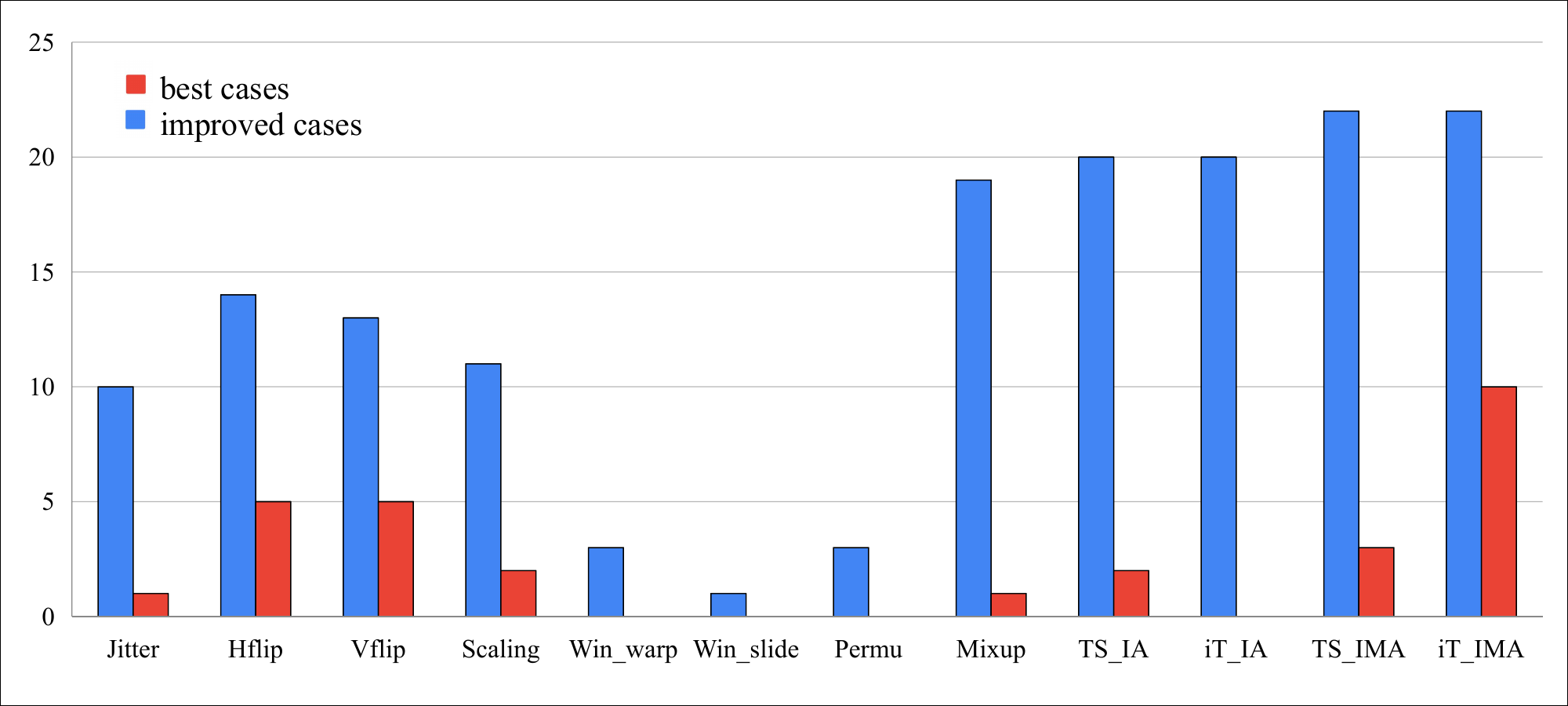} 
\caption{Comparison of the number of improvement cases and the best-case performance among eight augmentation methods, IA, and IMA on the ETT dataset.}
\label{fig5}
\end{figure}
\textbf{Results.}  
Table \ref{tab:result} demonstrates that Imputed-data Augmentation (IA) significantly improves performance, especially on the ETT dataset, achieving enhancements in 20 out of 24 cases, with notable success in all 8 instances using the iTransformer model (Fig. \ref{fig5}). Combining IA with Mixup (IMA) further strengthens results, improving 22 out of 24 cases on the ETT dataset, including 10 best-case outcomes. IMA also slightly outperforms Mixup alone on the Illness and Exchange Rate datasets. 

However, IA and IMA struggle in some scenarios, particularly with DLinear on the ETTm1 dataset. This is due to the DLinear model's simplified architecture, which cannot fully leverage complex temporal patterns or augmented data diversity, highlighting the need for advanced temporal feature extraction capabilities.
\\
\begin{figure}[h]
\centering
\includegraphics[scale=0.245]{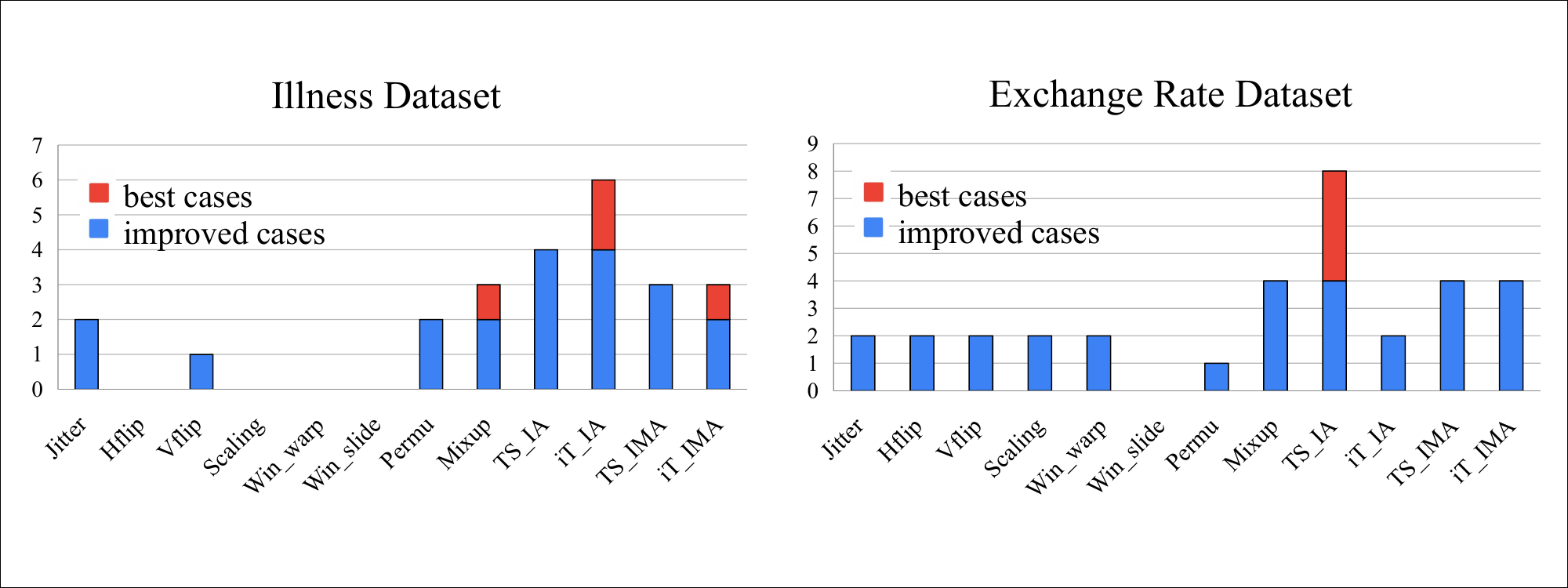} 
\caption{Comparison of the number of improvement cases and the best-case performance among eight augmentation methods, IA, and IMA on Illness, Exchange Rate datasets.}
\label{fig6}
\end{figure}
For the Illness and Exchange Rate datasets, IA achieves improvements in 4 out of 6 cases for both datasets, with peak performance in 2 Illness cases and 4 Exchange Rate cases (Fig. \ref{fig6}). These datasets' simplicity, characterized by many zero values, enables models like DLinear and iTransformer to effectively learn patterns without requiring significant augmentation, reducing the impact of augmentation methods. In contrast, TimesNet, with its convolutional operations, is more sensitive to augmentation, where IA outperforms IMA, suggesting that standalone imputation better suits the characteristics of these datasets.

In conclusion, IA and IMA demonstrate robust performance improvements across models and datasets. While IA occasionally surpasses IMA for specific datasets, the combined approach of IMA offers a versatile solution with more consistent performance across diverse scenarios, highlighting its advantage over existing methods.

\section{Conclusion}
In this study, we propose Imputation-based Mixup Augmentation (IMA), a method that enhances time series forecasting by leveraging SSL training to capture trends and patterns in the data while preserving essential characteristics. By combining Imputation with Mixup, IMA not only increases data diversity but also improves model generalization, leading to better forecasting performance. Our results demonstrate that this approach outperforms Mixup alone, highlighting its potential to generate more diverse and resilient training data. Moreover, IMA may not yield optimal results for every forecasting model and dataset, but it opens promising avenues for further exploration and development in this direction.
\bibliography{aaai25}

\end{document}